\documentclass[10pt, a4paper]{article}

\usepackage{lrec2026} 
\usepackage{booktabs} 
\usepackage{booktabs, multirow} 

\title{CzechDocs: A Multiway Parallel Dataset of Formatted Documents for Minority Languages in Czechia}

\name{Josef Jon, Ondřej Bojar} 

\address{Charles University, Faculty of Mathematics and Physics \\
Institute of Formal and Applied Linguistics \\
Prague, Czech Republic \\
  jon@ufal.mff.cuni.cz}

\abstract{
We present CzechDocs, a multiway parallel dataset of formatted documents (HTML, DOCX, and PDF) covering Czech and minority languages used in Czechia—primarily Ukrainian and English, with smaller portions of Vietnamese, Russian and other languages. The dataset is designed to support the evaluation of machine translation systems that aim to preserve document formatting during translation. We provide a comparison of the most common approaches to format-preserving machine translation on a validation subset of the dataset. This validation split, together with the evaluation toolkit, is publicly released for further research. A held-out test split will be reserved for a future shared task focused on document-level translation with formatting preservation.
 \\ \newline  }

\begin{document}

\maketitleabstract
\section{Introduction}

Despite progress in neural machine translation, the preservation of markup tags and document structure during translation continues to present challenges in practical deployment. Real-world translation workflows involve complex structured documents---web pages, software localization files, technical documentation and other content---where maintaining the integrity of HTML tags, XML structure and other markup elements is essential for downstream processing and publication.

This challenge has become particularly acute in scenarios requiring rapid deployment of multilingual content, such as crisis communication during humanitarian emergencies. The 2022 Russian invasion of Ukraine precipitated a refugee crisis, with over 500,000 Ukrainian citizens seeking temporary protection in the Czech Republic alone. This situation created an urgent need for high-quality translation of critical information---ranging from legal documents and government websites to educational materials and public health communications---while preserving the markup structure necessary for web publication and accessibility.

While machine translation evaluation has made significant strides in assessing lexical and semantic accuracy through metrics like BLEU, COMET, and human evaluation, comprehensive evaluation of production translation systems requires additional consideration of document structure preservation. We provide a dataset consisting of multiparallel formatted documents in Czech and one or more languages of minorities living in the Czech Republic.

We use our dataset to run a set of simple evaluations of the most common strategies to handle translation with markup. The emergence of LLMs as translation engines introduces new opportunities and questions in this context. In practical experience, at least some LLMs can handle translating the inputs with tags correctly. We compare keeping the tags in the input and prompting an LLM to carry them over to the translation to a more traditional detag-and-project approach, where tags are not shown to the MT model (or LLM, in our case) and are restored after the translation using word alignment. We also evaluate the impact of different prompting strategies---instructing the model to preserve tags, or providing no explicit markup-related guidance. We show effects of these strategies on automated metrics scores.

\subsection{Contributions}

This paper makes two contributions to the field of machine translation evaluation and markup-aware translation:

\textbf{1. A curated dataset for tag-aware translation evaluation.}
We present a curated parallel corpus of documents in DOCX, HTML and PDF formats, with domain mostly centered around help and instructions for foreigners who move into Czechia. Documents are drawn from authentic sources including government websites, legal forms, educational materials, and public health documentation\footnote{The dataset and the tools are available at \\ \url{https://github.com/cepin19/CzechDocs}}. 

\textbf{2. Comparison of tag-aware translation approaches.}
We evaluate multiple translation configurations spanning LLM-based translation, extarnal tag reinsertion heuristics, prompt engineering variations, alternative models, and context-aware translation.

\subsection{Paper organization}
The remainder of this paper is organized as follows:  Section~2 describes the dataset construction process. Section~3 details our evaluation framework and enumerates the translation approaches. Section~4 presents quantitative findings. Section~5 surveys existing research on markup-aware translation. Section~6 discusses concludes our work.

\section{Dataset}
\label{sec:dataset}

We present a curated multilingual parallel corpus specifically designed for evaluating machine translation systems on localization content with markup preservation. The dataset focuses on civic integration materials, information for foreigners moving to Czech Republic.

\subsection{Data Collection and Sources}

Our dataset comprises 316 total parallel language mutations of documents (77 unique documents) collected from 14 public Czech-language websites, mostly targeting foreign residents. Table~\ref{tab:document_origins} shows the distribution of documents by source website. The primary sources include:

\begin{itemize}
    \item \textbf{Educational portals}: Organization for teaching Czech language to foreigners (\textit{cestina-pro-cizince.cz}, 29 documents)
    \item \textbf{Information portals}: Expat-focused websites (\textit{newinprague.cz}, 20 documents) offering practical information about living in Prague, including guides, articles, and multimedia content.
    \item \textbf{Government services}: Official Czech government websites (\textit{mpsv.cz}, \textit{ipc.gov.cz}, \textit{mv.gov.cz}, \textit{mzv.gov.cz}, 16 documents total) providing employment regulations, immigration procedures, and official forms.
    \item \textbf{Municipal websites}: Local government portals (\textit{mestocernosice.cz}, \textit{kraj-jihocesky.cz}, \textit{kr-karlovarsky.cz}, 4 documents) offering local services and information for residents.
    \item \textbf{Public services}: Czech Ombudsman's office (\textit{ochrance.cz}, 2 documents), integration centers (\textit{icpraha.com}, 3 documents), and other civic organizations.
\end{itemize}

Data collection was performed through a combination of manual and automated web scraping. For HTML documents, we developed custom Python-based web scrapers using BeautifulSoup\footnote{\url{https://beautiful-soup-4.readthedocs.io/en/latest/}} to download parallel versions of web pages across multiple languages. For part of the PDF documents, we used targeted crawlers to identify and download parallel document pairs, matching Czech and Ukrainian versions by filename patterns and document structure. Other PDF and DOCX files were manually collected from official sources providing downloadable forms and templates. For machine translation experiments, we also converted all documents into XLIFF, Moses InlineText and plaintext formats.

\subsection{Document Formats and Alignment}

The dataset includes three original documents formats:

\begin{itemize}
    \item \textbf{HTML documents} (249 files, 79\%): These documents were \textit{manually edited and aligned at the segment level}, ensuring precise correspondence between source and target XLIFF segments. This manual alignment process involved inspecting extracted segments, identifying different or extra content in one of the languages and correcting the HTML source so that the resulting XLIFF segments are parallel. This level of alignment makes HTML documents suitable for fine-grained evaluation metrics.
    
    \item \textbf{PDF documents} (55 files, 17\%): Primarily official forms, informational brochures, and workplace safety materials from government agencies. PDF documents are \textit{aligned only at the document level} -- we have verified that corresponding PDFs are translations of each other, but individual segments within the documents are not aligned.
    
    \item \textbf{DOCX documents} (12 files, 4\%): Application forms, request templates, and administrative documents. Like PDFs, DOCX files are \textit{aligned at the document level only}. Each document is a complete translation of its parallel version, but internal XLIFF segment boundaries may differ between language versions.

\end{itemize}

\subsection{Preprocessing Pipeline}

All documents were processed through a pipeline to extract translatable content while preserving markup structure:

\begin{enumerate}
    \item \textbf{Cleanup and editing}: Some documents were not exact translations of each other, for example, they had extra content in one of the languages, some parts of the content were not translated (typically cookies disclaimer and similar) or the formatting was not the same across language versions. We manually edited the language mutations so that they are exact translations of each other, with the same formatting. 
    \item \textbf{Format conversion}: PDF files were first converted to DOCX using \texttt{pdf2docx}\footnote{\url{https://github.com/ArtifexSoftware/pdf2docx}}, which preserves paragraph boundaries, formatting, and basic document structure.
    
    \item \textbf{XLIFF extraction}: We used Okapi Tikal~\footnote{https://okapiframework.org/} localization toolkit to extract translatable content from HTML and DOCX documents into XLIFF, Moses Inline and plaintext formats. Tikal applies format-specific filters to identify translatable text while preserving inline markup as XLIFF tags.
    
    \item \textbf{Segmentation}: Okapi's default segmentation rules were applied to split content into translation units. For HTML documents, we manually verified segmentation boundaries to ensure consistent alignment across language pairs.

\end{enumerate}


\subsection{Segment Alignment}
\label{sec:alignment-tool}
For a \emph{subset} of HTML documents we performed manual segment-level alignment by editing the HTML source: we inspected extracted MOS/XLIFF segments, identified differing or extra content in one language, and corrected the HTML so that the resulting segments are parallel across language versions.

For \emph{all} HTML documents in the dataset we applied an alignment and reexport workflow so that every language version shares the same segment structure as the main (Czech) document. We've implemented a simple alignment tool that (1) loads Moses InlineText (MOS) segments extracted from HTML via Okapi Tikal, (2) displays them in an alignment grid where annotators can exclude segments, merge adjacent segments, edit segment text, or mark segments for per-language deletion, and (3) applies these edits to the MOS files. The segments in other languages can also be translated into the main language (Czech in our case) using MT so that the annotators can look at these translations during the alignment process.

The tool then \textbf{reexports} the edited MOS back into the original format (HTML or DOCX) by merging segment content into the \emph{structure} of the first language (Czech) document: for each language we run Tikal's merge command (\texttt{-lm}) using the Czech source file as the skeleton and the edited MOS for that language as the segment content. Thus every reexported document has the same segment count and boundaries as the Czech version, which ensures consistent alignment across all language mutations. This process yields a uniform, segment-aligned corpus suitable for sentence-level and tag-level evaluation while preserving the original formatting and markup structure. However, the result is less faithful to the original documents than the manual interventions in the source HTML files themselves, since untranslatable content (e.g. images) that was different in the original language mutations will be copied from the main language document after the reexport. Typical problematic example would be if a web page uses a country flag to show the currently selected language -- in that case, all the reexported mutations would show the main language flag. After a check of our reexported documents, we did not run into any similar issues.

\subsection{Dataset Statistics}

Table~\ref{tab:dataset_stats} presents comprehensive statistics for each language in the dataset. The corpus contains \textbf{60,153 translatable segments} across \textbf{15 languages}, with a total of \textbf{271,111 words} and \textbf{126,833 markup tags}. The dataset is dominated by Czech-Ukrainian parallel content (75 document pairs), reflecting the primary use case of materials for Ukrainian immigrants in the Czech Republic. Additional parallel content exists for Vietnamese, English, and Russian, with smaller collections for 10 other languages.

The average segment length is 4.5 words, characteristic of localization content which tends toward shorter, self-contained translation units (UI labels, form fields, navigation items). Notably, \textbf{90.3\% of segments contain at least one markup tag}, making this dataset particularly suitable for evaluating tag preservation capabilities. The high markup density (2.1 tags per segment on average) reflects the structured nature of web and document content.

Table~\ref{tab:language_pairs} shows the distribution of parallel documents by language pair. The dataset contains 76 Czech-Ukrainian parallel documents, 47 Czech-Vietnamese documents, and 47 Ukrainian-Vietnamese documents, among other combinations. This multilingual coverage enables both bilingual and multilingual evaluation scenarios.

\subsection{Dataset Split}

For evaluation purposes, we split the dataset into validation and test sets based on HTML documents (which have segment-level alignment). We randomly selected 19 HTML documents as a validation set, with the remaining documents reserved for future testing. The split maintains language distribution and includes documents from diverse sources. We publish the validation set upon publishing this paper, and we hold out the test set to organize a shared tasked for markup-aware MT.

\subsection{Availability}

The complete dataset, including source documents, XLIFF files, MOS format files, plaintext versions, and processing scripts, is available at \url{https://github.com/cepin19/CzechDocs} (excluding the held-out test set part, which will be released later). We release the dataset to support research in tag-aware machine translation and localization evaluation.

\begin{table}[ht]
\centering
\caption{Document Sources by Website}
\label{tab:document_origins}
\begin{tabular}{lr}
\toprule
Source Website & Documents \\
\midrule
cestina-pro-cizince.cz & 29 \\
newinprague.cz & 20 \\
mpsv.cz & 12 \\
icpraha.com & 3 \\
ipc.gov.cz & 2 \\
mestocernosice.cz & 2 \\
ochrance.cz & 2 \\
czechia.mfa.gov.ua & 1 \\
mv.gov.cz & 1 \\
mzv.gov.cz & 1 \\
edu.cz & 1 \\
kraj-jihocesky.cz & 1 \\
smocr.cz & 1 \\
kr-karlovarsky.cz & 1 \\
\bottomrule
\end{tabular}
\end{table}

\begin{table*}[ht]
\centering
\caption{Dataset Statistics by Language}
\label{tab:dataset_stats}
\begin{tabular}{lrrrrr}
\toprule
Language & Documents & Segments & Words & Tags & Avg Words/Seg \\
\midrule
UK & 75 & 14,252 & 63,784 & 27,848 & 4.5 \\
CS & 77 & 13,847 & 64,157 & 27,534 & 4.6 \\
VI & 47 & 10,087 & 46,815 & 23,063 & 4.6 \\
EN & 24 & 8,353 & 29,691 & 18,818 & 3.6 \\
RU & 22 & 8,107 & 26,004 & 18,874 & 3.2 \\
VN & 58 & 4,112 & 32,076 & 7,938 & 7.8 \\
FR & 3 & 334 & 2,643 & 182 & 7.9 \\
ES & 3 & 329 & 2,691 & 176 & 8.2 \\
SR & 1 & 141 & 481 & 70 & 3.4 \\
MN & 1 & 128 & 457 & 59 & 3.6 \\
AR & 1 & 123 & 218 & 16 & 1.8 \\
RO & 1 & 88 & 584 & 0 & 6.6 \\
BG & 1 & 87 & 463 & 0 & 5.3 \\
DE & 1 & 84 & 522 & 0 & 6.2 \\
PL & 1 & 81 & 525 & 0 & 6.5 \\
\midrule
\textbf{Total} & \textbf{316} & \textbf{60,153} & \textbf{271,111} & \textbf{124,578} & \textbf{4.5} \\
\bottomrule
\end{tabular}
\end{table*}

\begin{table}[ht]
\centering
\caption{Parallel Documents by Language Pair}
\label{tab:language_pairs}
\begin{tabular}{lr}
\toprule
Language Pair & Documents \\
\midrule
CS--UK & 75 \\
CS--VI & 47 \\
UK--VI & 47 \\
CS--VN & 31 \\
UK--VN & 31 \\
VI--VN & 27 \\
CS--EN & 24 \\
CS--RU & 22 \\
RU--UK & 22 \\
EN--RU & 22 \\
EN--UK & 22 \\
RU--VI & 20 \\
EN--VI & 20 \\
ES--FR & 3 \\
CS--ES & 3 \\
\bottomrule
\end{tabular}
\end{table}

\section{Methods}
There are multiple approaches to format-preserving MT. A very common method in localization workflows is to translate detagged text and then reinsert tags using word alignments (called also \textit{detag-and-project} approach). The typical pipeline involves converting documents (e.g., \texttt{.docx}, \texttt{.html}) to an intermediate representation such as XLIFF using tool like Okapi filters, segmenting text into units that preserve inline tags, translating the plain-text segments, optionally obtaining word alignments, and finally projecting tags from the source to the target via heuristic and/or alignment-based methods before reconstructing the original file format. An alternative approach is to present the full input to the model, including markup tags. To reduce complexity, tags are often replaced with single-token placeholders, which the models should to copy to the correct positions in the output. The models can also be finetuned on authentic or synthetic training data that contain markup to improve the efficiency of this method. Finally, recent research and practical experience has shown that large language models (LLMs) can often translate tagged segments directly—without specialized preprocessing or fine-tuning, provided that the prompt or instruction emphasizes correct tag preservation.
In this work, we ran a small-scale experiment to show the possible use of our dataset. We use an LLM for translation and we compare text translation and tag placement quality for:
\begin{itemize}
\item \textbf{Detag-and-project:} The model translates plain text only; tags are reinserted into the translation using word alignments from \texttt{awesome-align} \cite{dou2021word} using heuristics implemented in Charles Translator\footnote{\url{https://github.com/ufal/lindat-translation/tree/document_translation_llm}}.
\item \textbf{Direct tagged input:} The model receives full, tagged segments in Moses InlineText format with a simple translation prompt.
\item \textbf{Prompt with tag emphasis:} Same as (2), but the prompt explicitly instructs the model to preserve tag positions accurately.
\end{itemize}

The results of this comparison are presented in the following section.

\begin{table*}[!ht]\centering
\begin{tabular}{lrrrrrr}\toprule
\textbf{Model} &\textbf{Input} &\textbf{D\&P} &\textbf{Prompt} &\textbf{Tagged BLEU} &\textbf{Untagged BLEU} \\\midrule
\multirow{5}{*}{\textbf{gpt-4.1-nano}} &Plaintext &No &Normal &0.9 &55.7 \\
&MOS &No &Normal &25.4 &54.6 \\
&MOS &No &Explicit &87.8 &55.4 \\
&MOS &No &Explicit+Context &79.5 &54.6 \\
&Plaintext &Yes &Normal &89.0 &50.4 \\ \midrule
\multirow{5}{*}{\textbf{aya-expanse-8B}} &Plaintext &No &Normal &0.8 &51.8 \\
&MOS &No &Normal &75.7 &52.4 \\
&MOS &No &Explicit &83.6 &51.4 \\
&Plaintext &Yes &Normal & 89.6 & 49.5 \\
\bottomrule
\end{tabular}
\caption{BLEU scores with and without tags on the validation set. The second column shows the input format, i.e. how was the source presented to the translation engine. The third column shows if detag-and-project approach was used. Third column shows the prompt type. }\label{tab:results}
\end{table*}
\section{Evaluation results}

We compare two LLMs: \textit{Aya-expanse-8b} from CohereForAI and \textit{gpt-4.1-nano} from OpenAI. We translated the validation parts of our dataset by the methods described above. We measure BLEU score including the tags (relative to tagged reference) and BLEU of detagged, plain text (relative to untagged reference). The results are shown in Table \ref{tab:results}.

We used 2 different input formats, Moses InlineText, presented as MOS in the table and plaintext with the tags stripped. The \textit{detag-and-project} column shows whether the tags were restored in the target after translation using word alignment.
We used 3 different prompts: \textit{normal}, which only tells the model to translate the input segment, \textit{explicit} which explicitly states that the markup should be transfered to the target and \textit{explicit+context}, which in addition to the current segment contains the rest of the document source segments and their translations up to the current segment for context.

We see a slightly better performance of the \textit{gpt-4.1-nano}. In our settings, the model proved sensitive to prompt changes, explicit instructions to preserve markup led to noticeable improvements. For \textit{aya-expanse-8b}, explicit prompt performed better as well, but the difference was smaller. On average, the \textit{detag-and-project} and tag-aware prompting approaches achieved comparable results, but with greater variance across individual documents. Some documents favored one approach markedly over the other. We leave the identification of document properties that cause one method to outperform the other for future research.

\section{Related Work}

Handling markup and formatting in machine translation (MT) has long been a practical challenge in localization workflows involving HTML, XML, or other inline-coded formats. Early systematic treatments, such as \citet{muller-2017-treatment,joanis-etal-2013-transferring}, provide overview of markup processing in SMT systems, comparing strategies such as masking, segmentation-based reinsertion, and alignment-based transfer. Although developed for SMT, many of these principles remain relevant in the NMT era.

\citet{zenkel-etal-2021-automatic-bilingual} formalize the bilingual markup transfer task, i.e. the situation where source with the markup and translation without markup is available and the task is to transfer the markup from the source to the target. They provide code and metrics for alignment-based tag projection. \citet{hanneman-dinu-2020-markup} introduce a data augmentation approach to improve NMT with markup. \citet{ryu-etal-2022-data} expands on this idea by introducing synthetic markup pairs and phrase-table-inspired augmentation to increase tag consistency. 

Recent studies leverage large language models (LLMs) for this task. \citet{dabre-etal-2023-study} evaluate zero- and few-shot prompting for markup transfer, demonstrating that LLMs can handle structured tag propagation without alignment information, although results vary by language pair and tag complexity.

The \textit{Salesforce Localization XML} dataset \citep{hashimoto-etal-2019-high} remains one of the few public benchmarks for markup-aware MT. It includes parallel data between English and sixteen target languages, along with evaluation utilities for BLEU, tag accuracy, and placement metrics. They also compare multiple approaches to tag translation on this dataset. Similarly, the WMT20 tagged test suites \citep{hanneman-dinu-2020-markup} provide controlled sentence-level evaluation data for markup translation. \citet{buschbeck-etal-2022-multilingual} present a multilingual multiway structured translation dataset for Asian languages.

 On the tooling side, several open-source projects, such as \texttt{m4loc}\footnote{\url{https://github.com/achimr/m4loc}}, the \textit{Charles Translator}\footnote{\url{https://github.com/kukas/document-translation}}, and \texttt{LibreTranslate}\footnote{\url{https://github.com/LibreTranslate/LibreTranslate}}, implement variations of alignment-based reinsertion. Public evaluation resources such as Lilt’s \texttt{markup-transfer} toolkit\footnote{\url{https://github.com/lilt/markup-tag-evaluation}} and Amazon’s \texttt{mt-markup-tags}\footnote{\url{https://github.com/amazon-science/mt-markup-tags}} offer open evaluation code for measuring structural fidelity alongside translation quality.

\section{Conclusions}
We created a dataset for evaluation of format-preserving MT for Czech and multiple languages used by minorities and expats in Czech Republic. We show a possible use-case by running an evaluation of two approaches to markup-aware translation using LLMs. The results show that while the LLMs we evaluated are capable of transferring the markup, having specialized implementation of detagging and reinsertion based on word alignment slightly improved our automated scores on tagged text. On the other hand, the untagged BLEU was slightly lower for the detag-and-project approach. The validation split of the dataset is released at the same time as this paper, while we hold out a larger test split to be used in a future shared task on the topic. 
\section{Acknowledgements}
This work was partially supported by SVV project number 260 821, by Czech Ministry of Education, Youth and Sports (grant MŠMT OP JAK Mezisektorová spolupráce CZ.02.01.01/00/23\_020/0008518) and by National Recovery Plan funded project MPO 60273/24/21300/21000 CEDMO 2.0 NPO. 

It has been using language resources and tools developed and/or stored and/or distributed by the LINDAT/CLARIAH-CZ project of the Ministry of Education, Youth and Sports of the Czech Republic (project LM2023062).

\section{Bibliographical References}\label{sec:reference}
\bibliographystyle{lrec2026-natbib}
\bibliography{lrec2026-example}

\bibliographystylelanguageresource{lrec2026-natbib}
\bibliographylanguageresource{languageresource}

\end{document}